\documentclass[conference]{IEEEtran}
\IEEEoverridecommandlockouts

\usepackage{cite}
\usepackage{amsmath,amssymb,amsfonts}
\usepackage{algorithmic}
\usepackage{graphicx}
\usepackage{booktabs}
\usepackage{multirow}
\usepackage{stfloats}
\usepackage{textcomp}
\usepackage{xcolor}
\usepackage[hidelinks]{hyperref}
\def\BibTeX{{\rm B\kern-.05em{\sc i\kern-.025em b}\kern-.08em
    T\kern-.1667em\lower.7ex\hbox{E}\kern-.125emX}}
\begin{document}

\title{Graph-Based Personalized Memory for LLM Agents: Representation, Evolution, Retrieval, and Evaluation
\thanks{\textsuperscript{†}Corresponding author.}
}

\author{\IEEEauthorblockN{Dac Duy Anh Nguyen}
\IEEEauthorblockA{\textit{School of ICT} \\
\textit{Griffith University}\\
Brisbane, Australia \\
rodo.nguyen@griffith.edu.au}
\and
\IEEEauthorblockN{Zhangchi Qiu}
\IEEEauthorblockA{\textit{School of ICT} \\
\textit{Griffith University}\\
Gold Coast, Australia  \\
z.qiu@griffith.edu.au}
\and
\IEEEauthorblockN{Shigeng Chen}
\IEEEauthorblockA{\textit{School of ICT} \\
\textit{Griffith University}\\
Gold Coast, Australia  \\
shigeng.chen@griffithuni.edu.au}
\and
\IEEEauthorblockN{Alan Wee-Chung Liew\textsuperscript{†}}
\IEEEauthorblockA{\textit{School of ICT} \\
\textit{Griffith University}\\
Gold Coast, Australia  \\
a.liew@griffith.edu.au}
}

\maketitle

\begin{abstract}
Large Language Model (LLM) agents are evolving from single-session tools toward long-term personal assistants that must adapt to individual users across tasks, contexts, and interactions. This shift makes memory a core requirement for personalization, since user preferences, goals, constraints, relationships, and past experiences are accumulated gradually and often change over time. Graph-based personalized memory provides a structured way to model such user information through explicit relations, temporal context, and evidence links. Such representations can model not only what an agent remembers about a user but also how memories are connected, revised, and retrieved to support personalized decisions. However, existing work remains fragmented across personalized agents and generic graph memory frameworks, making it difficult to understand the design space as a whole. This survey develops a lifecycle-oriented view of graph-based personalized memory for LLM agents. We organize existing studies around memory representation, memory evolution, memory retrieval, and memory evaluation. We further compare key design choices, discuss current evaluation practices, and open challenges in building reliable long-term personalized agents. This survey aims to clarify how graph-based memory can support adaptive, controllable, and user-centric LLM agents.
The repository for this survey is continuously maintained at \url{https://github.com/icedpanda/awesome-personalized-graph-memory}.
\end{abstract}


\section{Introduction}
\label{sec:introduction}

Recent advances in large language models (LLMs) have enabled a new class of autonomous agents that can follow complex instructions, plan over multiple steps, use different tools, and carry out long-horizon tasks in dynamic environments~\cite{luo2025large, xuPersonalizedGenerationLarge2025}. Building on these capabilities, they are beginning to serve as personal assistants that reflect and shape their own behaviors to individual users, such as OpenClaw~\cite{OpenClaw2026} and Hermes~\cite{HermesAgent2026}.
As agents move toward sustained, user-facing interaction, personalization becomes a central focus, since effective long-term assistance requires not only general reasoning but also an understanding of the user's goals, preferences, constraints, and prior experiences~\cite{xuPersonalizedLLMPoweredAgents2026, qiuReasoningUserPreferences2025,qiu2025knowledge,yangGraphbasedAgentMemory2026}.

Despite LLMs' increasingly large context windows, including a user's full interaction history in every prompt remains impractical~\cite{jiangPersonaMemv2PersonalizedIntelligence2025, huangLiCoMemoryLightweightCognitive2026}.
Personalized agents therefore require a persistent memory that can accumulate user-specific knowledge across sessions, update it as preferences and circumstances change, and ultimately guide agent behavior tailored to individual users~\cite{xuHierarchicalLongTermSemantic2026, xuPersonalizedLLMPoweredAgents2026}.

Early agent memory systems explored a range of formats to retain user-specific knowledge across sessions, from dialogue-derived records~\cite{yuan-etal-2025-personalized, packer2024memgptllmsoperatingsystems} to segment-level or compressed natural-language memories~\cite{pan2025Secom, nan2025nemori}. These designs can record and recall individual facts but typically treat each entry as self-contained, offering limited capacity to capture how facts interrelate, how they evolve over time, or how one event gives rise to the next~\cite{yangGraphbasedAgentMemory2026, huangLiCoMemoryLightweightCognitive2026}.

To overcome these limitations, graph-based memory has emerged as a structured alternative that organizes user facts, preferences, interaction events, and the relations among them into an interconnected memory space~\cite{prestonrasmussenZepTemporalKnowledge2025, liangPersonaAgentGraphRAGCommunityAware2025, maoBiMemBidirectionalConstruction2026, yangGraphbasedAgentMemory2026}, transitioning agent memory from a passive archive of entries into an explicit model of the user that preserves how personal information is connected and how it changes over time.

This design is a natural fit for personalization because user-specific memory often connects people, places, items, tasks, goals, domains, episodes, and time~\cite{xuUserMemoryReasoning2020, wangCraftingPersonalizedAgents2024, xuPersonalizedLLMPoweredAgents2026}. Unlike flat logs, summaries, or vector stores, graph-based memory can express this structure directly thanks to its intrinsic ability to link scattered evidence into coherent user personas, to traverse multi-hop context that resolves who and what a request refers to, to trace each profile claim back to its supporting interactions, and to track how preferences and circumstances evolve~\cite{maoBiMemBidirectionalConstruction2026, phamvanMemORAIMemoryOrganization2026, zhaofenwuGAMHierarchicalGraphbased2026, xuHierarchicalLongTermSemantic2026, qiu-gcrs2025,luo2026greasoner}.

Such capabilities turn the graph from a passive store of user facts into an operational model of the user, although realizing them depends on how the graph is constructed, updated, and queried rather than on the adoption of a graph alone~\cite{yangGraphbasedAgentMemory2026, liuGraphAugmentedLargeLanguage2025}.

Although recent surveys have examined personalized agents and agent memory, they use different primary lenses. Surveys of personalized agents frame personalization through agent capabilities, including profile modeling, memory, planning, and action execution. They treat persistent user memory as one component, rather than as the central object of representation~\cite{xuPersonalizedLLMPoweredAgents2026}. The survey of graph-based agent memory organizes the field around memory extraction, storage, retrieval, and evolution across broad agent settings. It considers personalization alongside other applications~\cite{yangGraphbasedAgentMemory2026}. Broader agent-memory surveys focus on memory-module design and evaluation, modular architectures and benchmarked strategies, or an evolutionary progression from storage to reflection and experience~\cite{zhangSurveyMemoryMechanism2024, wu2026memoryllmeramodular, StoragetoExperience}. Together, these perspectives do not provide a unified account of how graph structures represent, update, retrieve, and evaluate persistent personalized memory for LLM agents.

\begin{figure}[t]
    \vspace{-0.1cm}
    \centering
    \includegraphics[width=0.96\linewidth]{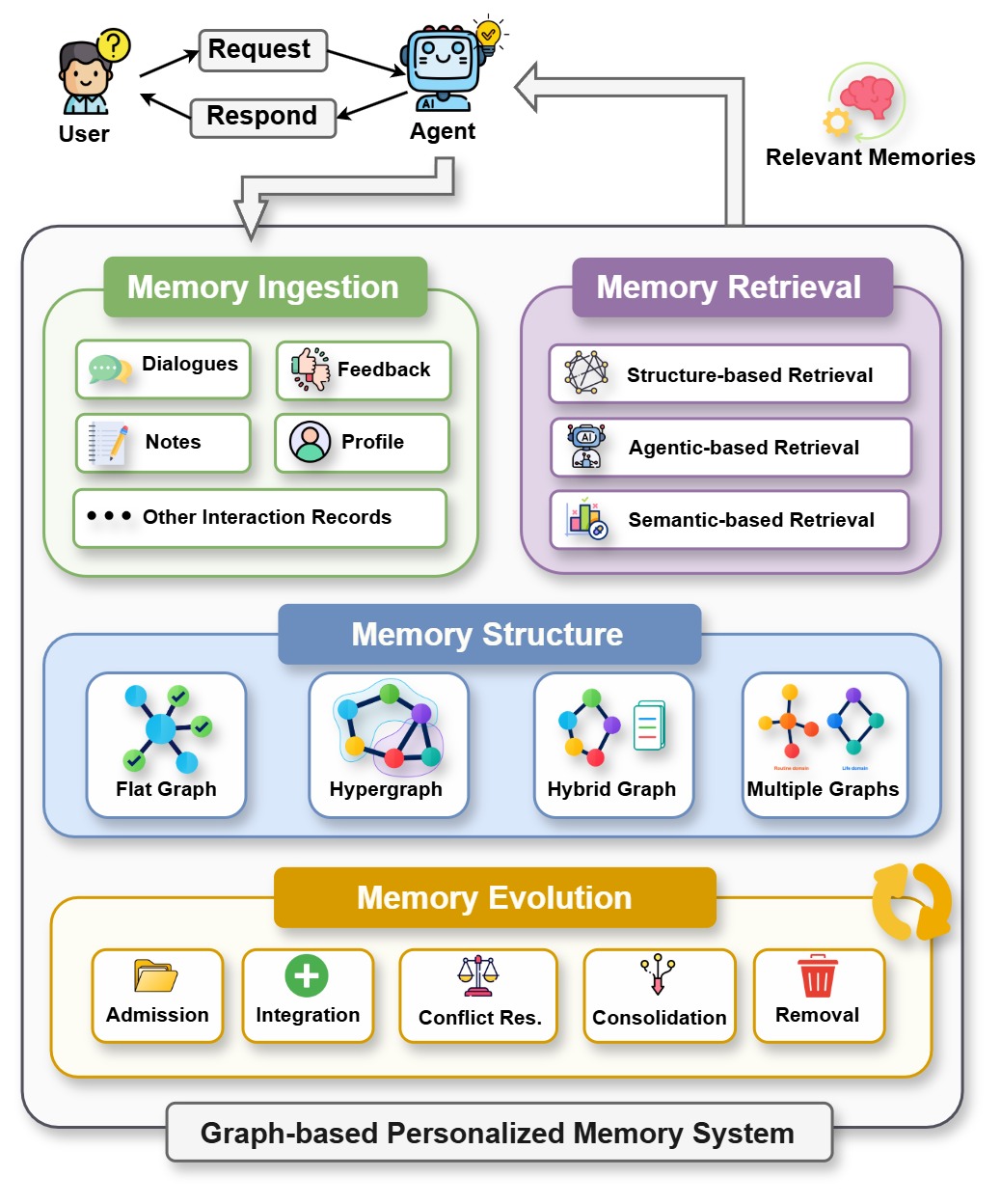}
    \caption{Overview of a graph-based personalized agent memory system.}
    \label{fig:overview}
    \vspace{-0.3cm}
\end{figure}

To address this gap, this survey develops a lifecycle-oriented view of \textbf{graph-based personalized memory} for LLM agents. We synthesize how existing systems construct user-centered graph representations, update and consolidate new evidence, retrieve personalized context for response grounding, and evaluate memory quality through current benchmarks and metrics. This perspective connects graph design choices with the practical requirements of long-term personalization and identifies open challenges in personalized memory. As Fig.~\ref{fig:overview} illustrates, graph-based personalized memory connects memory ingestion, structure, evolution, and retrieval within the personalized agent workflow.

Our contributions are as follows:
\begin{itemize}
    \item \textbf{Taxonomy}. We develop a lifecycle taxonomy that organizes existing studies on graph-based personalized memory around representation, evolution, retrieval, and evaluation.
    \item \textbf{Evaluation}. We summarize existing benchmarks and metrics and identify gaps in graph quality, user control, lifecycle validity, and realistic personalization.
    \item \textbf{Timely Review}. For each category, we review recent methods and discuss their motivations, graph structures, and user-modeling functions.
    \item \textbf{Future Directions}. We outline several open research directions to guide future research on this promising topic.
\end{itemize}
\section{Preliminaries}
\label{sec:preliminaries}

\begin{table*}[t]
    \centering
    \caption{Dimensions of personal memory.}
    \label{tab:personal-memory-dimensions}
    \small
    \renewcommand{\arraystretch}{1.25}
    \begin{tabular}{@{}lcc@{}}
        \toprule
        Dimension & What it Captures & Example \\
        \midrule
        \textbf{Semantic} & Stable facts, preferences, goals, or constraints & ``I'm vegetarian; learning French'' \\
        \textbf{Temporal} & When a memory holds and how it changes & ``Joined the gym last month'' \\
        \textbf{Relational} & Links to people, places, tasks, or items & ``Likes the coffee shop downtown'' \\
        \textbf{Experiential} & Past events, outcomes, routines, and habits & ``Missed my morning flight last time'' \\
        \textbf{Affective} & Emotional valence and intensity (satisfaction, frustration) & ``I hated that noisy hotel'' \\
        \bottomrule
    \end{tabular}
    \vspace{-0.2cm}
\end{table*}

\subsection{Graph-based Personalized Memory}
Personalization refers to an LLM agent adapting its responses, retrieval, and actions to an individual user. This adaptation depends on \textbf{graph-based personalized memory}, which contains user-specific state covering a user's facts, preferences, goals, constraints, relationships, and experiences. The memory persists across sessions, is stored outside the model parameters, and is organized as an explicit graph. The \emph{memory} aspect separates it from the transient prompt context, and the \emph{personalized} aspect separates it from general knowledge shared across users. We therefore exclude sole generic graph-augmented generation and agent memory unless they support user modeling.

Agent memory has traditionally been stored as raw dialogue logs~\cite{yuan-etal-2025-personalized,packer2024memgptllmsoperatingsystems} or natural-language summaries~\cite{pan2025Secom,nan2025nemori}, which recall individual entries but leave their relations, supporting evidence, validity, and contradictions implicit. Graph-based memory instead represents user information through \emph{nodes} (the remembered items) and \emph{edges} (the relations among them), making these connections and revisions explicit and queryable. The rest of this survey follows this representation through its lifecycle: how it is constructed (Section~\ref{sec:representation}), evolves (Section~\ref{sec:evolution}), is retrieved (Section~\ref{sec:retrieval}), and is evaluated (Section~\ref{sec:evaluation}).

\subsection{Personal Memory Dimensions}
\label{subsec:personal-memory-dimensions}

Personal memory is the user-specific state an agent preserves beyond the current context window, spanning facts, preferences, goals, constraints, episodes, feedback, emotions, and corrections to earlier claims. Rather than a flat checklist of traits, we characterize it along five \emph{dimensions} that capture what kind of information is remembered, summarized in Table~\ref{tab:personal-memory-dimensions}. The dimensions are not mutually exclusive: a single memory may be semantic in content, bounded in time, relationally anchored, and affectively charged at once. Together, these definitions define what a graph must represent; later sections show how graph structures encode, update, and retrieve it.

\section{Memory Representation}
\label{sec:representation}

We examine graph memory representation at the element and structure levels. At the \emph{element level}, systems specify node roles, relation types, and lifecycle metadata. At the \emph{structure level}, systems adopt one or more structural designs, including flat, hierarchical, hypergraph, hybrid, and multi-graph designs. These choices together shape evidence traceability, state validity, and retrievable granularity.

\subsection{Memory Elements}
\label{subsec:nodes-edges}

We distinguish four functional node roles, which may overlap within one graph. \emph{Evidence nodes} preserve original interaction records, such as turns, episodes~\cite{GAAMAGraphAugmented2026}, or sentences~\cite{wuSGMemSentenceGraph2025}. \emph{Fact nodes} store extracted atomic assertions about the user or related entities~\cite{chhikaraMem0BuildingProductionReady2025,phamvanMemORAIMemoryOrganization2026}. \emph{Abstraction nodes} synthesize repeated observations into higher-level user models, such as personas, scenes~\cite{maoBiMemBidirectionalConstruction2026}, or reflections~\cite{GAAMAGraphAugmented2026}. \emph{Domain-specific nodes} capture entities relevant to particular topics, such as personal interests, jobs, or more specific topics like health care and patient routines~\cite{songDEMENTIAPLANAgentBasedFramework2025}. A node may serve more than one role.

Common relation functions include temporal, provenance, associative, and domain-specific links. \emph{Temporal relations} encode event order or progression~\cite{GAAMAGraphAugmented2026,zhaofenwuGAMHierarchicalGraphbased2026}. \emph{Provenance relations} connect assertions or abstractions to source evidence~\cite{phamvanMemORAIMemoryOrganization2026,huangLiCoMemoryLightweightCognitive2026}. \emph{Associative relations} link contextually or semantically related memories~\cite{wuSGMemSentenceGraph2025,jiMemoryReconstructedNot2026}. \emph{Domain-specific relations} encode task-relevant semantics~\cite{cuiLOOMPersonalizedLearning2025,songDEMENTIAPLANAgentBasedFramework2025}. Beyond relations, lifecycle metadata such as timestamps, validity intervals, and invalidation markers~\cite{chhikaraMem0BuildingProductionReady2025,prestonrasmussenZepTemporalKnowledge2025} can distinguish active user state from historical evidence. Evidence-rich designs preserve source context but increase storage demand; fact- or abstraction-centered designs offer compactness at the risk of propagating extraction errors; domain-specific schemas improve task alignment but reduce generality.

\subsection{Memory Structures}
\label{subsec:memory-structures}

\begin{figure}
    \centering
    \resizebox{\linewidth}{!}{\includegraphics{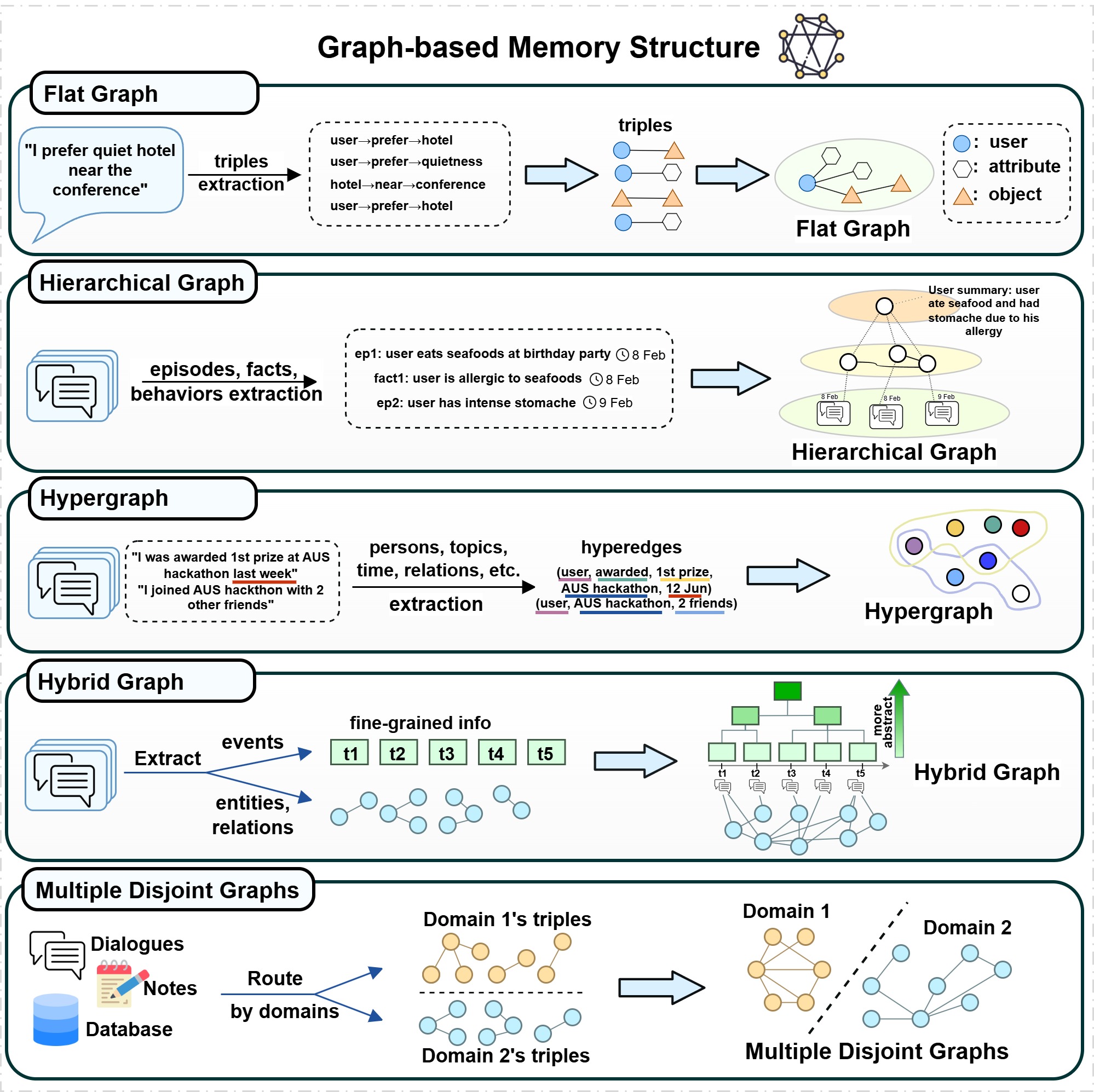}}
    \caption{Representative graph-based memory patterns for personalized LLM agents.}
    \label{fig:structures}
    \vspace{-0.2cm}
\end{figure}

We organize the systems into five representation patterns, as shown in Fig.~\ref{fig:structures}: flat, hierarchical, hypergraph, hybrid, and multiple disjoint graphs. These patterns are not mutually exclusive, and a system may combine several patterns. Flat, hierarchical, and hypergraph patterns describe organization within one graph. Hybrid designs combine graphs with non-graph memory substrates, whereas multiple-graph designs partition memory across separate graph instances.

\subsubsection{Flat Graph}
We use \emph{flat graph} to refer to a memory graph that stores user-related facts, preferences, and relations as nodes and typed edges without explicit abstraction layers. Heterogeneous node and edge types do not themselves constitute a hierarchy. UMGR~\cite{xuUserMemoryReasoning2020} is a domain-oriented heterogeneous graph linking users, items, attributes, and opinions for recommendation; KGT~\cite{sunKnowledgeGraphTuning2024} stores personalized factual triples representing editable user beliefs; Mem0\textsuperscript{g}~\cite{chhikaraMem0BuildingProductionReady2025} constructs an open-domain entity-relation graph incrementally from conversation. These systems share the absence of explicit abstraction layering while their node schemas and scopes differ. Flat topology stores user-related facts and relations explicitly but does not itself separate source evidence from extracted facts or from higher-level abstractions. Flat graphs can still include provenance and validity metadata.

\subsubsection{Hierarchical Graph}
Hierarchical graph memories organize user evidence into multiple layers, making user state accessible at different levels of specificity. Hierarchy need not be reduced to monotonic abstraction; cross-level links can keep evidence reachable from higher-level representations~\cite{GAAMAGraphAugmented2026,maoBiMemBidirectionalConstruction2026}. Hierarchies serve two main functions in the systems considered here. For evidence-to-abstraction organization, GAAMA~\cite{GAAMAGraphAugmented2026} preserves turns as episode nodes, extracts facts and concepts at an intermediate level, and synthesizes reflections capturing higher-order user patterns; Bi-Mem~\cite{maoBiMemBidirectionalConstruction2026} groups fact-level memories into thematic scenes and distills them into a global persona, with cross-level links enabling calibration against local evidence. For granularity-preserving organization, GAM~\cite{zhaofenwuGAMHierarchicalGraphbased2026} organizes active event graphs, topic-level nodes, and archived events into distinct layers, while SGMem~\cite{wuSGMemSentenceGraph2025} decomposes conversations into sessions, turns, and sentences, keeping fine-grained detail accessible alongside coarser summaries and extracted facts.

Hierarchical organization supports traceability across levels between raw evidence and higher-level user representations. Generated abstractions such as personas or reflections may propagate extraction errors or overgeneralize transient evidence. Systems that preserve bidirectional or cross-level links can check higher-level abstractions against lower-level evidence, but this introduces cross-layer consistency requirements that grow with depth.

\subsubsection{Hypergraph}
Hypergraphs preserve joint context whose meaning depends on several connected elements. HingeMem~\cite{zhongHingeMemBoundaryGuided2026} groups boundary-triggered dialogue segments into hyperedges indexed by elements such as participants, locations, and times; HyperMem~\cite{yueHyperMemHypergraphMemory2026} organizes memory into topic, episode, and fact levels, using episode hyperedges to group episodes under the same topic and fact hyperedges to group facts within an episode. Both preserve joint contextual groupings, but HingeMem emphasizes matching a query to an event, whereas HyperMem emphasizes grouping related information across a long conversation before coarse-to-fine (top-down) retrieval.

\subsubsection{Hybrid Graph}
Hybrid memory systems distribute the user model across graph topology and a complementary non-graph substrate, such as a tree, summary store, passage store, or evidence buffer. The graph provides relational access among user facts, entities, and events; the non-graph substrate holds long-form interaction evidence, temporal organization, or source text outside the graph. H-Mem~\cite{yuHMemNovelMemory2026} combines a temporal-semantic tree that organizes memory fragments from timestamped events to higher-level summaries with an entity graph that records relations and links entities to the fragments that mention them.

Other hybrid designs use a lightweight graph as an index to richer stores. LiCoMemory~\cite{huangLiCoMemoryLightweightCognitive2026} stores entity-relation pairs in a graph, while session summaries and dialogue chunks carry the content. MemWeaver~\cite{yeMemWeaverWeavingHybrid2026} partitions memory into a structured relation graph, an experience store for repeated interaction patterns, and a passage store for source text. These specialized stores separate relational indexing from evidence storage but require consistency across components.

\subsubsection{Multiple Disjoint Graphs}
Multiple-graph designs split user memory across separate graph instances that serve distinct, specialized functions. Evidence for this design is currently limited to DEMENTIA-PLAN~\cite{songDEMENTIAPLANAgentBasedFramework2025}, which maintains a daily routine graph for time-sensitive care context and a life memory graph for longer-term autobiographical identity. This arrangement separates user states by timescale and purpose across distinct instances. The cited work does not establish whether these graph instances are formally disjoint, and further evidence is needed to assess the generality of this design. Nevertheless, this pattern may be useful for highly specific use cases where memory can be stored and retrieved more efficiently because its location is known in advance.

\subsection{Discussion}
\label{subsec:representation-discussion}

Representation determines which user evidence a system retains and which dependencies later lifecycle operations can use. Evidence-rich designs preserve provenance and support correction, while fact- and abstraction-centered designs provide compact user models but may discard context or propagate extraction errors. Hierarchical and multi-store representations also require consistency across levels or substrates. Lifecycle metadata can distinguish an active state from historical evidence, but its presence does not ensure that updates propagate correctly. Thus, representation is best understood as a substrate for memory evolution and retrieval rather than as a guarantee of personalization quality.
\section{Memory Evolution}
\label{sec:evolution}

\begin{figure}[t]
    \centering
    \includegraphics[width=0.95\linewidth]{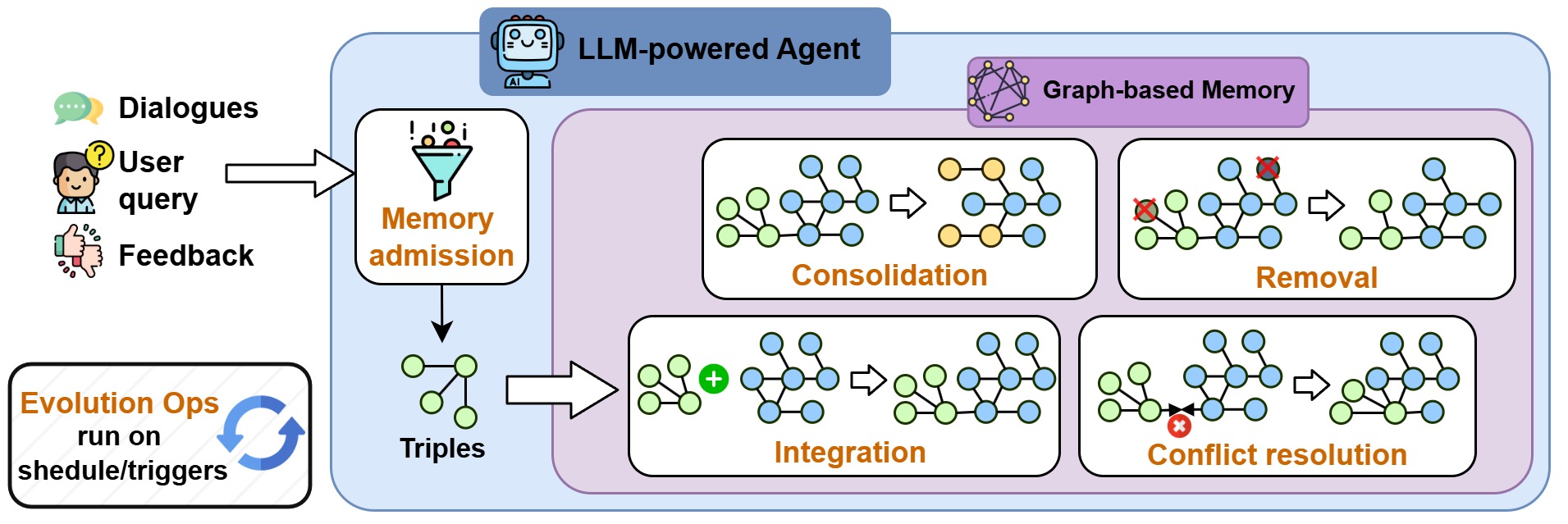}
    \caption{Evolution of graph-based personalized memory.}
    \label{fig:evolution}
    \vspace{-0.2cm}
\end{figure}

Memory evolution maintains a usable active user model while preserving historical evidence where needed. As new evidence arrives, the system decides whether to admit it, integrate it into the existing structure, resolve conflicts with prior states, consolidate accumulated memory, and remove outdated records. This section examines these five evolution operations (illustrated in Fig.~\ref{fig:evolution}).

\subsection{Memory Admission}
\label{subsec:triggers-admission}

Admission decides whether incoming evidence becomes persistent memory and at what functional type or granularity. In KGT~\cite{sunKnowledgeGraphTuning2024}, user feedback is treated as a trigger to add or replace factual triples in its graph. MemORAI~\cite{phamvanMemORAIMemoryOrganization2026} applies selective filtering and compression to retain user-persona-relevant episodic content, facts, and preferences while compressing less informative discourse. MemGuard~\cite{haMemGuardPreventingMemory2026} assigns each memory a functional role (e.g., fact, procedural instruction, constraint) at write time and preserves boundaries among functionally distinct memory types. Admission protocol is important as over-admission risks embedding transient claims as durable user state, while under-admission can discard evidence that later interactions require.

\subsection{Memory Integration}
\label{subsec:local-integration}

After admission, integration attaches evidence to the existing graph structure at a local write scale. Append-and-link preserves provenance to the originating turn, session, or chunk, as in LiCoMemory~\cite{huangLiCoMemoryLightweightCognitive2026}. To reduce redundancy when admitting memory, Mem0\textsuperscript{g}~\cite{chhikaraMem0BuildingProductionReady2025} reuses semantically matched nodes and invalidates obsolete edges, while HingeMem~\cite{zhongHingeMemBoundaryGuided2026} consolidates near-duplicate boundary hyperedges. GAM~\cite{zhaofenwuGAMHierarchicalGraphbased2026} uses buffered attachment, writing new utterances into an active event graph before later consolidation into its global graph. This ensures GAM only consolidates semantically complete units and reduces semantic drift in the main graph.

\subsection{Conflict Resolution}
\label{subsec:conflict-resolution}

Conflict resolution addresses incoming evidence that changes the validity or scope of a stored record. Three distinct conflict types apply: (1) temporal supersession, where a newer state replaces an earlier one; (2) user correction, where stored facts or inferences are identified as wrong; and (3) contextual coexistence, where apparently contradictory states remain valid in different contexts. These types should not be collapsed into a single newest-wins rule. 

AriadneMem~\cite{zhuAriadneMemThreadingMaze2026} handles temporal conflicts by merging static duplicates and preserving state transitions as temporal edges. More generally, conflict handling can overwrite an earlier value, mark it inactive while retaining its evidence as in GRAVITY~\cite{sunGRAVITYArchitectureAgnosticStructured2026}, or preserve successive states as queryable versions like ZEP~\cite{prestonrasmussenZepTemporalKnowledge2025}.

\subsection{Consolidation}
\label{subsec:consolidation-abstraction}

Consolidation reorganizes accumulated memory after multiple writes. GAM~\cite{zhaofenwuGAMHierarchicalGraphbased2026} promotes completed event buffers into a topic-level structure while archiving detailed event graphs, serving abstraction and archival functions. All-Mem~\cite{lvAllMemAgenticLifelong2026} applies gated SPLIT, MERGE, and UPDATE topology edits while preserving immutable evidence, serving deduplication, restructuring, and archival functions. Consolidation therefore serves abstraction, deduplication, restructuring, and archival, although individual systems implement different subsets. When observations are promoted into higher-level abstractions, errors in source evidence may persist into derived representations, and provenance links can support later correction without guaranteeing it.

\subsection{Memory Removal}
\label{subsec:memory-removal}

Memory removal takes two distinct forms: forgetting and deleting. Forgetting reduces a memory's exposure, resolution, weight, or retrieval likelihood while some representation remains stored. ScrapMem~\cite{changScrapMemBioinspiredFramework2026} applies Optical Forgetting, which progressively reduces the resolution of older scrapbook pages, suppressing low-value detail while lowering storage cost. 

Deletion changes what remains stored, active, or recoverable. Kumiho~\cite{parkGraphNativeCognitiveMemory2026} uses versioned belief revision: immutable revisions are retained, mutable tags indicate the active state, and deprecated memories are excluded from normal retrieval unless explicitly requested. This design retains full progression history for auditing. 

Removing source evidence can require revising or removing derived facts, summaries, personas, or higher-level abstractions produced from it to prevent the active user model from becoming internally inconsistent: a profile claim may remain active after the interaction supporting it has been removed. The reliability and generality of dependency propagation across personalized memory systems remain open evaluation questions.

\subsection{Discussion}
\label{subsec:evolution-discussion}

Implemented systems connect evolution operations across different timescales. For example, GAM~\cite{zhaofenwuGAMHierarchicalGraphbased2026} buffers events before consolidation, and All-Mem~\cite{lvAllMemAgenticLifelong2026} periodically edits topology while retaining immutable evidence. AriadneMem~\cite{zhuAriadneMemThreadingMaze2026} records temporal state transitions while Kumiho~\cite{parkGraphNativeCognitiveMemory2026} separates current and historical revisions through mutable tags. These implementations link incoming evidence, structural reorganization, and active-state selection, but they do not establish one universally preferable lifecycle design.

\section{Memory Retrieval}
\label{sec:retrieval}

Retrieval maps a current request to a bounded subset of the user's persistent memory graph. We group retrieval methods into three mechanisms that may be combined: (1) similarity-based retrieval provides initial candidate generation or graph anchoring; (2) structure-based retrieval expands evidence through relations, abstraction levels, or relational views; and (3) adaptive and agentic retrieval controls query interpretation, expansion, or compression according to query intent or intermediate evidence. Their shared objective is to recover the user-specific evidence needed for the current response without exposing unrelated personal history.

\subsection{Similarity-Based Retrieval}
\label{subsec:similarity-retrieval}

Similarity-based retrieval covers lexical, keyword, and dense-vector matching over memory stores. In graph memory, it commonly generates initial candidates or maps a query to graph anchors~\cite{wu2026memoryllmeramodular}. EMG-RAG~\cite{wangCraftingPersonalizedAgents2024} applies this selection problem to an editable graph of smartphone memories, retrieving user-specific records for downstream personalized tasks. Flat top-$k$ selection can return correlated or near-duplicate interaction fragments~\cite{huRAGAgentMemory2026}, while semantic proximity alone may miss evidence connected through temporal, causal, entity, or abstraction relations.

\subsection{Structure-Based Retrieval}
\label{subsec:structure-retrieval}

Structure-based retrieval uses graph organization to expand or refine initial candidates. For personalized memory, this expansion connects a current request to user-specific evidence distributed across relations, abstraction levels, or temporal views. GAM~\cite{zhaofenwuGAMHierarchicalGraphbased2026} first retrieves topic nodes and then accesses archived event graphs when finer detail is required. MAGMA~\cite{jiangMAGMAMultiGraphBased2026} represents each memory item across semantic, temporal, causal, and entity graphs and selects the view that best matches query intent.

Associative links provide another retrieval signal. AssoMem~\cite{zhangAssoMemScalableMemory2025} anchors dialogue utterances to automatically extracted clues in an associative memory graph and combines relevance, importance, and temporal alignment for candidate ranking. xMemory~\cite{huRAGAgentMemory2026} first selects complementary groups and memory components, then expands to source segments or messages when additional evidence is needed. Structural retrieval can recover user evidence distributed across sessions or abstraction levels, but uncontrolled expansion adds irrelevant nodes, unsupported paths, and context cost.

\subsection{Adaptive and Agentic Retrieval}
\label{subsec:adaptive-retrieval}

Adaptive and agentic retrieval changes query interpretation, graph exploration, stopping, routing, or compression according to the request and intermediate evidence. In personalized memory, this control helps distinguish the current user state from historical evidence and limits retrieval to the detail needed by the present request. MRAgent~\cite{jiMemoryReconstructedNot2026} uses a Cue-Tag-Content graph to iteratively explore and prune retrieval paths according to intermediate evidence, adapting memory access during inference.

Adaptive control also governs routing and compression. PRISM~\cite{pengPRISMParetoEfficientRetrieval2026} combines intent-sensitive path selection with evidence compression to retain high-utility spans under a context budget. HingeMem~\cite{zhongHingeMemBoundaryGuided2026} controls both which element-indexed routes to activate and how deeply to retrieve according to query demands. APEX-MEM~\cite{banerjeeAPEXMEMAgenticSemiStructured2026} resolves conflicting or evolving information at retrieval time through a multi-tool retrieval agent.

\subsection{Discussion}
\label{subsec:retrieval-discussion}

These three mechanisms can be combined in a retrieval pipeline: similarity anchoring identifies candidate entry points; structural expansion recovers distributed user evidence across relations, levels, or views; and adaptive control selects or compresses results under a context window.  

Graph structure does not automatically improve retrieval since controlled comparisons show that results can depend on foundational system settings beyond graph structure~\cite{huDoesMemoryNeed2026}. Graph retrieval is therefore useful only when the representation exposes meaningful user relations, evolution maintains their validity, and retrieval returns relevant evidence.

\section{Evaluation}
\label{sec:evaluation}

Evaluation in personalized memory is still mostly downstream: benchmarks typically test whether a memory system answers correctly, retrieves supporting evidence, verifies updated user state, or reduces context cost over long histories. Since each benchmark targets different memory capabilities, Table~\ref{tab:benchmark-coverage} summarizes representative personalized-memory benchmarks and their evaluation metrics, adapting the metric categories discussed in~\cite{xuPersonalizedLLMPoweredAgents2026}.

The dominant benchmark family still tests long-horizon conversational recall. LoCoMo~\cite{maharanaEvaluatingVeryLongTerm2024} is a common evaluation benchmark for graph-memory papers, including production-ready memory with a graph variant~\cite{chhikaraMem0BuildingProductionReady2025}, topology-aware graph reasoning~\cite{zhuAriadneMemThreadingMaze2026}, learned traversal~\cite{dongmingjiangHAGEHarnessingAgentic2026}, multi-scale user memory~\cite{tianRGMemRenormalizationGroupinspired2025}, and graph-native versioned memory~\cite{parkGraphNativeCognitiveMemory2026}. LongMemEval~\cite{wuLongMemEvalBenchmarkingChat2024} evaluates longer histories, knowledge updates, single-session preference questions, temporal reasoning, and abstention. These benchmarks are valuable because they force retrieval under multi-session noise, but they mostly observe whether the final answer is correct rather than whether the memory graph itself is correct.

Personalization-specific benchmarks shift the target from factual recall to user-state modeling. PersonaMem~\cite{jiangKnowMeRespond2025} tests dynamic user-profile and preference tracking, while PersonaMem-v2~\cite{jiangPersonaMemv2PersonalizedIntelligence2025} scales this setting to implicit personas, longer contexts, larger preference coverage, and agentic memory and therefore puts personalized memory systems to a new challenge. RealMem~\cite{bianRealMemBenchmarkingLLMs2026} extends personalization into project-oriented scenarios where goals, schedules, and dependencies evolve across sessions. PerLTQA~\cite{duPerLTQAPersonalLongTerm2024} is useful because it organizes personal long-term memory into semantic memory, including profiles and social relationships, and episodic memory, including events and dialogues.

A smaller set of benchmarks tests requirements that are especially important for graph-based memory. EngramaBench~\cite{acunaEngramaBenchEvaluatingLongTerm2026} isolates whether structured graph retrieval helps across persona spaces, temporal traces, and cross-space associations. EvoMemBench~\cite{wangEvoMemBenchBenchmarkingAgent2026} evaluates self-evolving memory across in-episode/cross-episode and knowledge/execution settings, while EvoArena~\cite{xuEvoArenaTrackingMemory2026} stresses persistent environment change and preference-evolution chains. ActMemEval~\cite{zhangActMemBridgingGap2026} tests whether retrieved memory supports implicit constraint and counterfactual reasoning, while ATM-Bench~\cite{meiAccordingMeLongTerm2026} adds multimodal, multi-source personal evidence. StructMemEval~\cite{shutovaEvaluatingMemoryStructure2026} asks whether memory is organized correctly.

Across these benchmarks, reported metrics fall into five groups:
\begin{itemize}
    \item \textbf{Answer quality:} F1, exact match, multiple-choice accuracy, task success, and LLM-as-judge score.
    \item \textbf{Evidence retrieval:} Recall@k, NDCG@k, answer-turn retrieval accuracy, evidence-location accuracy, and multimodal evidence recall.
    \item \textbf{User-state modeling:} preference-category accuracy, profile or persona consistency, and correctness or helpfulness of personalized responses.
    \item \textbf{Memory evolution and structure:} update success, step or chain accuracy under changing environments, forgetting/update outcomes, structure-requiring task accuracy, and failure-attribution accuracy.
    \item \textbf{Efficiency:} retrieved-context size, token usage, latency, construction cost, query-time cost, and storage reduction.
\end{itemize}

\begin{table*}[t]
\centering
\caption{Representative personalized-memory benchmarks and evaluation signals. \(\checkmark\): explicit; \(\triangle\): partial/indirect. Ans.: answer quality; Ret.: evidence retrieval; User: user-state modeling; Evol.: evolution over time; MemOrg.: memory organization; Cost: efficiency; MM: multimodal evidence.}
\label{tab:benchmark-coverage}
\scriptsize
\setlength{\tabcolsep}{3pt}
\renewcommand{\arraystretch}{1.10}
\begin{tabular}{@{}p{0.20\textwidth}p{0.20\textwidth}p{0.20\textwidth}ccccccc@{}}
\toprule
Benchmark & Focus & Setting & Ans. & Ret. & User & Evol. & MemOrg. & Cost & MM \\
\midrule
\multicolumn{10}{@{}l}{\textit{Long-horizon recall and retrieval}} \\
\midrule
LoCoMo~\cite{maharanaEvaluatingVeryLongTerm2024} & Long conversational memory & Real text-image conversations & \(\checkmark\) & \(\checkmark\) & \(\triangle\) & -- & -- & -- & \(\checkmark\) \\
LongMemEval~\cite{wuLongMemEvalBenchmarkingChat2024} & Long interactive memory & Long chat histories & \(\checkmark\) & \(\checkmark\) & \(\triangle\) & \(\triangle\) & -- & \(\triangle\) & -- \\
\midrule
\multicolumn{10}{@{}l}{\textit{Personalization and user-state modeling}} \\
\midrule
PersonaMem~\cite{jiangKnowMeRespond2025} & Dynamic user profiling & Simulated chats & \(\checkmark\) & -- & \(\checkmark\) & \(\checkmark\) & -- & -- & -- \\
PersonaMem-v2~\cite{jiangPersonaMemv2PersonalizedIntelligence2025} & Implicit persona inference & Long persona chats & \(\checkmark\) & -- & \(\checkmark\) & \(\checkmark\) & -- & \(\checkmark\) & \(\checkmark\) \\
RealMem~\cite{bianRealMemBenchmarkingLLMs2026} & Project-oriented memory & Multi-session projects & \(\checkmark\) & \(\checkmark\) & \(\checkmark\) & \(\checkmark\) & -- & \(\checkmark\) & -- \\
PerLTQA~\cite{duPerLTQAPersonalLongTerm2024} & Personal long-term QA & Semantic/episodic memory & \(\checkmark\) & \(\checkmark\) & \(\checkmark\) & -- & -- & -- & -- \\
\midrule
\multicolumn{10}{@{}l}{\textit{Graph structure and evolution}} \\
\midrule
EngramaBench~\cite{acunaEngramaBenchEvaluatingLongTerm2026} & Graph-vs-context memory & Structured retrieval & \(\checkmark\) & \(\checkmark\) & \(\triangle\) & -- & \(\checkmark\) & \(\checkmark\) & -- \\
EvoMemBench~\cite{wangEvoMemBenchBenchmarkingAgent2026} & Self-evolving memory & Episodic tasks & \(\checkmark\) & -- & -- & \(\checkmark\) & -- & \(\checkmark\) & -- \\
EvoArena~\cite{xuEvoArenaTrackingMemory2026} & Environment evolution & Persistent environments & \(\checkmark\) & -- & \(\checkmark\) & \(\checkmark\) & -- & \(\checkmark\) & \(\triangle\) \\
ActMemEval~\cite{zhangActMemBridgingGap2026} & Action memory & Constraint reasoning & \(\checkmark\) & \(\checkmark\) & \(\triangle\) & -- & \(\triangle\) &  \(\checkmark\) & -- \\
ATM-Bench~\cite{meiAccordingMeLongTerm2026} & Multimodal personal evidence & Referential QA & \(\checkmark\) & \(\checkmark\) & \(\checkmark\) & \(\checkmark\) & -- & \(\triangle\) & \(\checkmark\) \\
StructMemEval~\cite{shutovaEvaluatingMemoryStructure2026} & Memory structure & Structure-requiring tasks & \(\checkmark\) & \(\triangle\) & -- & -- & \(\checkmark\) & -- & -- \\

\bottomrule
\end{tabular}
\end{table*}

\section{Challenges and Future Directions}
\label{sec:future-directions}

Personalized graph memory remains a young but promising research frontier. Its central challenge is not merely storing user information in graph form but building a long-term, trustworthy, and actionable model of an individual user, which in turn motivates several future directions.

\noindent\textbf{Scalable Lifelong Personal Memory.} Personalized agents should support years of interaction rather than isolated sessions or short benchmark histories. Future work should build lifelong graph memories that grow with the user, consolidate recurring evidence, preserve historical context, adapt to preference drift, and at the same time, maintain accurate and efficient retrieval as memory grows. Achieving this may require new memory paradigms or architectures~\cite{quekMeMoMemoryModel2026}.

\noindent\textbf{Multimodal Personal Memory Graphs.} Current systems mostly construct memory from text, yet real personalization is grounded in multimodal evidence, including speech, images, documents, locations, actions, and interaction traces. Future work should develop graph memories that connect these heterogeneous signals while preserving their source modality, reliability, and temporal context.

\noindent\textbf{Causal and Counterfactual User Modeling.} Most existing memories are associative, recording what the user said or did. Stronger personalization requires understanding why a preference holds, when it changes, and what would happen under different contexts. Future graph memories should support causal and counterfactual reasoning over user state, distinguishing stable traits from temporary constraints or accidental observations.

\bibliographystyle{IEEEtran}
\bibliography{references}

\end{document}